\documentclass[letterpaper]{article} %

\PassOptionsToPackage{sort}{natbib}

\usepackage{aaai25}  %

\nocopyright

\usepackage{times}  %
\usepackage{helvet}  %
\usepackage{courier}  %
\usepackage[hyphens]{url}  %
\usepackage{graphicx} %

\usepackage{tabularx}
\usepackage{booktabs}
\usepackage{array}
\usepackage{siunitx}

\usepackage{xcolor}

\usepackage{natbib}  %
\usepackage{caption} %
\usepackage{algorithm}
\usepackage{algpseudocode}
\usepackage{amsmath}

\usepackage{newfloat}
\usepackage{listings}

\usepackage{amsmath}        %
\usepackage{algorithm}      %
\usepackage{algpseudocode}  %

\usepackage{enumitem}

\DeclareCaptionStyle{ruled}{labelfont=normalfont,labelsep=colon,strut=off} %
\floatstyle{ruled}
\newfloat{listing}{tb}{lst}{}
\floatname{listing}{Listing}
\usepackage{bibentry}

\usepackage{amssymb}
\usepackage[skip=3pt]{caption}

\title{AQA: Adaptive Question Answering in a Society of LLMs\\ via Contextual Multi-Armed Bandit}
\author{
Mohanna Hoveyda\textsuperscript{\rm 1}, 
Arjen P. de Vries\textsuperscript{\rm 1}, 
Maarten de Rijke\textsuperscript{\rm 2}, 
Harrie Oosterhuis\textsuperscript{\rm 1}, 
Faegheh Hasibi\textsuperscript{\rm 1}
}
\affiliations{
    \textsuperscript{\rm 1}Radboud University, Nijmegen, The Netherlands\\
\textsuperscript{\rm 2}University of Amsterdam, Amsterdam, The Netherlands\\
    \{mohanna.hoveyda, arjen.devries, harrie.oosterhuis, faegheh.hasibi@ru.nl\}@ru.nl, m.derijke@uva.nl
}

\begin{document}

\maketitle

\begin{abstract}
In question answering (QA), different questions can be effectively addressed with different answering strategies. Some require a simple lookup, while others need complex, multi-step reasoning to be answered adequately.
This observation motivates the development of a dynamic method that adaptively selects the most suitable QA strategy for each question, enabling more efficient and effective systems capable of addressing a broader range of question types.
To this aim, we build on recent advances in the orchestration of multiple large language models (LLMs) and formulate adaptive QA as a dynamic orchestration challenge. We define this as a contextual multi-armed bandit problem, where the context is defined by the characteristics of the incoming question and the action space consists of potential communication graph configurations among the LLM agents. We then train a linear upper confidence bound model to learn an optimal mapping between different question types and their corresponding optimal multi-LLM communication graph representation. 
Our experiments show that the proposed solution is viable for adaptive orchestration of a QA system with multiple modules, as it combines the superior performance of more complex strategies while avoiding their costs when simpler strategies suffice.%

\end{abstract}

\section{Introduction}

Large language models (LLMs) have facilitated the development of diverse question answering (QA) systems and pipelines, each with distinct performance across domains \citep{cuconasu2024power, li2024chainofknowledge, DBLP:conf/www/XuPSCC24, ram2023context, DBLP:journals/corr/abs-2403-14403}.
The increasing complexity of these QA pipelines stems from the integration of various steps, each designed to either mitigate particular errors introduced by a module (or interactions between them) or to address questions of different types or with varying requirements 
\citep{DBLP:journals/corr/abs-2403-14403, DBLP:conf/acl/TrivediBKS23}. 
Although these complex modular designs aim to enhance overall system robustness and accuracy in generating answers, it substantially increases inference costs.
Additionally, a single sophisticated answering strategy may not be the most suitable solution for all types of questions \citep{DBLP:journals/corr/abs-2403-14403}.

\subsubsection{Complex LLM-based QA systems.}
The large variety of recent retrieval augmented generation (RAG) approaches~\citep{ram2023context} provides a good example of the increasing complexity of QA systems.
RAG enables the use of external knowledge during inference without re-training or modifications to the LLM-architecture.
To improve its effectiveness, many additional parameters and modules have been proposed for RAG, e.g., retrieving subgraphs from a structured knowledge base alongside or instead of passages \citep{DBLP:journals/corr/abs-2402-11541, DBLP:conf/emnlp/BehnamGhaderMR23, DBLP:journals/tgdk/PanRKSCDJO0LBMB23, DBLP:journals/tkde/PanLWCWW24}, employing summarization techniques \citep{DBLP:journals/corr/abs-2404-16130}, introducing noise to retrieval results \citep{cuconasu2024power}, and natural language inference modules that preprocess the retrieved content for the LLM \citep{DBLP:conf/iclr/YoranWRB24}.
Another example of complex QA systems are those designed to cater to more complex questions, such as multi-hop questions~\citep{DBLP:conf/emnlp/Yang0ZBCSM18}, which require the integration of several pieces of knowledge. %
In such contexts, techniques have been proposed to enhance the deductive capabilities of LLMs, including CoT \citep{wei2022chain}, ToT \citep{DBLP:conf/nips/YaoYZS00N23}, LINC \citep{Olausson_2023}, and interleaved retrieval CoT \citep{DBLP:conf/acl/TrivediBKS23}.
Various recent studies on LLM orchestration propose to automate the integration of multiple steps in a QA system pipeline~\citep{DBLP:conf/nips/LiHIKG23, DBLP:journals/corr/abs-2305-17066, DBLP:journals/corr/abs-2402-16823, DBLP:conf/uist/ParkOCMLB23}.
However, complex approaches to QA systems do not always improve performance since LLMs can struggle to effectively use the expanded context \citep{DBLP:journals/tacl/LiuLHPBPL24}. %

\subsubsection{Adaptive QA Systems.}

In real-world use cases, not all types of questions that a QA system receives require the same amount of computation or the same sophisticated answering strategy. 
Accordingly, the complexity of building a reliable LLM-based QA system has increased, since it requires  effective collaboration between different modules, such as LLMs, IR modules, and any intermediate processing steps~\citep{li2024chainofknowledge, DBLP:conf/www/XuPSCC24}.
The rise in module count also greatly increases inference time and computational costs. %
Consequently, there is a growing emphasis on creating adaptive QA frameworks that dynamically adjust to varying question characteristics \citep{DBLP:conf/acl/MallenAZDKH23, DBLP:conf/iclr/AsaiWWSH24, DBLP:journals/corr/abs-2403-14403}.

\subsubsection{Approach.}

In this paper, we build on the recently-introduced concept of graph-based LLM orchestration to develop an adaptive QA system.
\citet{DBLP:journals/corr/abs-2402-16823} approach the problem of orchestrating multiple LLMs as a graph optimization problem, where each LLM-based module is defined as a sub-graph (or a node) and communication among them is formulated as edges in the final graph. 
We propose a novel framework for the adaptive selection and execution of graph configurations that are best-suited for the complexity level of each given question. 
We frame the adaptive orchestration of collaboration in multi-agentic systems as a contextual multi-armed bandit (CMAB) problem where the decision points are incoming questions and their characteristics serve as context to the CMAB, the action space consists of the set of all possible graph configurations.
In other words, the CMAB chooses which pipeline to apply per question, and thus, adapts its complexity and inference costs dynamically.
We evaluate our proposed adaptive orchestration framework in a QA setup.

\subsubsection{Contributions.} Our main contributions are the following: 
\begin{itemize}[leftmargin=*]
    \item We propose an \textit{Adaptive Question Answering} (AQA) framework for the adaptive orchestration of multi-agent LLM-based QA systems,
    that enables the execution of the most suitable answering strategy based per question.
    \item Our framework uniquely frames the orchestration of multi-agent LLM-based systems as a \emph{contextual multi-armed bandit} (CMAB) problem, motivated by the high effectiveness of CMABs in limited action spaces.
    This sets it apart from the policy gradient methods in previous work.
    \item To the best of our knowledge, our adaptive orchestration framework AQA is the first to optimize a combination of effectiveness and inference cost, enabling it to balance efficiency and performance when selecting answering strategies over varying question complexities. 
    \item We provide an experimental evaluation of the AQA framework over a multi-complexity-level QA dataset.
    Our results indicate that AQA successfully recognizes and applies an optimal mapping between question types and the most-suitable answering strategies. 
    
\end{itemize}

\section{Related Work}
\subsection{Adaptive Retrieval Augmented Generation}
Adaptive-Retrieval for RAG was first addressed by \citet{DBLP:conf/acl/MallenAZDKH23}, who propose a dynamic model that decides whether to retrieve external documents per incoming question.
Specifically, \citet{DBLP:conf/acl/MallenAZDKH23} propose a binary threshold-based framework that tries to distinguish whether questions contain popular or long-tail entities, and decides whether to retrieve accordingly. 
Inspired by reinforcement learning reward models, \citet{DBLP:conf/iclr/AsaiWWSH24} introduced Self-RAG, a language model that generates text interleaved with pre-embedded reflection tokens to critique and guide the generation process in real-time. 
\citet{DBLP:journals/corr/abs-2403-14403} propose Adaptive-RAG, a language model trained as a complexity classifier to predict question complexity, and subsequently, select the most suitable model accordingly.
Our framework is similar, but instead of assuming the optimal strategy for each complexity label, a CB is applied to learn which strategy works best.
Thereby, we also avoid relying on training language models for decision making, in contrast with previous work~\citep{DBLP:conf/iclr/AsaiWWSH24, DBLP:journals/corr/abs-2403-14403}, which can be prone to the hallucinations and unfaithful reasoning \citep{siegel-etal-2024-probabilities} that is common in LLMs.

\subsection{Multi-Agentic LLM Orchestration}
Several studies have tried to create an effective framework for letting multiple LLMs and related modules communicate to solve tasks \citep{DBLP:conf/iclr/HongZCZCWZWYLZR24, DBLP:journals/corr/abs-2308-08155}.
Most existing work focuses on orchestration without further optimization of the structure of these agents \citep{DBLP:journals/corr/abs-2308-08155, DBLP:journals/corr/abs-2305-17066}.
Inspired by Minsky's society of minds (SoM) \citep{minsky1988society}, which describes how smaller parts of a system can collaborate to achieve a goal, \citet{DBLP:journals/corr/abs-2305-17066} suggest a shift from relying on optimizing a single model for solving a task to the optimization of information flow between two or more models. %
This approach reveals a wide variety of possible orchestrations of (agentic) models, from which a selection can be made to match the task at hand.

\citet{DBLP:journals/corr/abs-2402-16823} propose GPTSwarm to optimize a society of language models.
GPTSwarm structures the QA system pipeline as a graph, and every computational operation, e.g., querying a large language model with a prompt, is represented as a node within that graph.
The dual-level optimization approach of GPTSwarm optimizes prompts at the node level while also enhancing the flow of information by pruning out edges and nodes that are not found useful.

Similar to \citet{DBLP:journals/corr/abs-2402-16823}, our orchestration framework is focused on graph optimization to improve the communication between modules.
However, \citet{DBLP:journals/corr/abs-2402-16823} apply policy gradients and produce a single graph structure for all questions, whereas we use a CMAB approach that can adapt the graph per individual question.

\section{Method: The Novel Adaptive Question Answering
Framework}

We frame \emph{adaptive question answering} (AQA) as a contextual multi-armed bandit (CMAB) problem \citep{langford2007epoch, MAL-068}.
Accordingly, our objective is to train a CMAB to select the answering strategy for each incoming question that optimizes effectiveness and efficiency.

\subsection{Question Context Vector}

Various features of a question can be important when choosing answering strategies, e.g., semantic information about the content of the question, or more miscellaneous features such as the length of its text, its knowledge domain, etc.
Let $Q_t$ denote an incoming question at timestep $t$; we represent $Q_t$ by a \textit{context vector} $x_t \in \mathbb{R}^d$ with dimensionality $d$ where each element $x_{ti}$ represents a feature of $Q_t$:
\[
x_t = 
[x_{t1}, x_{t2},  \ldots,  x_{td}].
\]
The context vector $x_t$ is all the information on which the CMAB bases its decision at timestep $t$.
For this work, we assume that the question-complexity of $Q_t$ can be inferred from $x_t$.

\begin{algorithm}[t]
\caption{Adaptive question answering via a contextual multi-armed bandit algorithm for multi-agent LLM orchestration.}
\label{alg:linucb}
\small  %
\begin{algorithmic}[1]
\State \textbf{Input:} the set of agents $\mathbf{L}$, exploration parameter $\alpha$  \label{alg:line:setofagents}
\State $\mathbf{V} \gets \emptyset$ \label{alg:line:startgraphconstruction}
\For{each agent $l_i \in \mathbf{L}$}
    \State $v_i \gets \text{Node}(l_i)$
    \State $\mathbf{V} \gets V \cup \{v_i\}$
\EndFor
\State $v_{\text{final}} \gets \text{MajorityVoting}()$
\State $\mathbf{V} \gets \mathbf{V} \cup \{v_\text{final}\}$

\State $\mathbf{E} \gets \{(v_i, v_j) \mid v_i, v_j \in V \land i \neq j \land v_i \not= v_\text{final}\}$
\State $\mathbf{G} \gets (\mathbf{V}, \mathbf{E})$ 
\State $\mathcal{G} \gets \{G_i \mid E(G_i) \subseteq E(G) \land \text{DAG}(G_i) \land deg^-(v_\text{final}) > 0$  \label{alg:line:endgraphconstruction}
\Statex \qquad\qquad\qquad $\land \; (\forall v_i \in V, \; deg(v_i) = 0 \lor \text{Path}(v_i, v_\text{final}))\}$ 

\For{each action $a \in \mathcal{G}$} \label{alg:line:startUCBinitialization}
    \State $\mathbf{A}_a \gets \mathbf{I}_d$ 
    \State $\mathbf{b}_a \gets \mathbf{0}$  \label{alg:line:endUCBinitialization}
\EndFor
\Statex 
\For{each timestep $t$} \label{alg:line:startmainloop}
    \State Wait for next question $Q_t$ and observe $x_t \in \mathbb{R}^d$ \label{alg:line:observequery}
    \For{each action $a \in \mathcal{G}$} \label{alg:line:startactionselection}
        \State $\theta_a \gets \mathbf{A}_a^{-1} \mathbf{b}_a$ %
        \State $p_{t,a} \gets \theta_a^T x_t + \alpha \sqrt{x_t^T \mathbf{A}_a^{-1} x_t}$ %
    \EndFor 
    \State $a_t \gets \arg\max_a p_{t,a}$ \label{alg:line:endactionselection} %
    \State Apply answering strategy $a_t$ and observe performance $\text{P}_t$ \label{alg:line:observeperformance}
    \State $r_t \gets \beta \cdot \text{P}_t - \gamma \cdot \text{T}_t$ \label{alg:line:computereward} %
    \State $\mathbf{A}_{a_t} \gets \mathbf{A}_{a_t} + x_t x_t^T$ \label{alg:line:startupdateparams}
    \State $\mathbf{b}_{a_t} \gets \mathbf{b}_{a_t} + r_t x_t$ \label{alg:line:endupdateparams}
\EndFor \label{alg:line:endmainloop}
\end{algorithmic}
\end{algorithm}
\subsection{Agents}
\label{subsec:Agents}
Let $\mathbf{L} = \{l_1, l_2, \ldots, l_n\}$ be the set of agents employed to generate an answer to a given question.
$L$ can include a variety of LLMs, both standard and augmented with advanced capabilities, e.g., retrieval \citep{ram2023context} or reasoning \citep{wei2022chain, DBLP:conf/acl/TrivediBKS23}, in addition to other auxiliary modules that feature a natural language interface \citep{DBLP:journals/corr/abs-2305-17066} such as query rewriters \citep{chan2024rq}, NLI models \citep{yoran2024making}, and theorem provers \citep{Olausson_2023}. 
While our framework works for any arbitrary $\mathbf{L}$, the experiments in this work follow the setup proposed by \citet{DBLP:conf/acl/TrivediBKS23} and used by \citet{DBLP:journals/corr/abs-2403-14403} and concern the following three agents:
\begin{equation}
\mathbf{L}' = \{ \textit{NoR}, \textit{OneR}, \textit{IRCoT} \}.
\end{equation}
Each of these agents are designed to handle questions of specific complexity levels.
We describe them in increasing order of their targeted complexity level:
\begin{itemize}[leftmargin=*]
\item \textbf{NoR}: An LLM with \textit{\textbf{no} \textbf{r}etrieval} augmentation. NoR answers questions without additional external data, making it most suited for simple non-knowledge-intensive questions that match its encoded parametric knowledge.
\item \textbf{OneR}: An LLM augmented by \textit{\textbf{one}-time \textbf{r}etrieval}.
Upon receiving a question, an IR module is called \emph{once} to provide $n$ potentially relevant documents alongside the question for the LLM.
OneR is most suited for questions that require access to external knowledge. 
\item \textbf{IRCoT}: An LLM augmented with retrieval and prompted to apply chain-of-thought (COT) reasoning.
The retrieved documents are interleaved in several back-and-forth reasoning steps. %
As the most complex and time-consuming strategy, IRCoT is most-suited for questions where NoR and OneR are expected to fail, i.e., ones that demand extensive knowledge or a synthesis of multiple pieces of knowledge to produce an accurate answer.
\end{itemize}

\subsection{Graph Design}
\label{subsection:collab}
We define a graph as $\mathbf{G= (V, E)}$, where $\mathbf{V}$ represents the set of nodes and $\mathbf{E}$ the set of directed edges.
Following the experimental setup outlined in \citep{DBLP:journals/corr/abs-2402-16823} for the MMLU dataset \citep{hendryckstest2021}, we choose $\mathbf{V}$ to be $\mathbf{L}$ with the addition of a final decision node $v_\text{final}$:
\begin{equation}
    \mathbf{V} = \{l_1, l_2, \ldots, l_n, v_{\text{final}}\}.
\end{equation}
The node $v_\text{final}$  aggregates the outputs of connected nodes through a majority vote, so that the most frequent answer from the agents becomes the final response.\footnote{See \citep{DBLP:journals/corr/abs-2402-16823} for other possible final decision nodes.}
The edges $\mathbf{E}$ represent the interactions between nodes, and, thus, are defined according to what interactions are possible between the specific agents.
To guarantee the graph is executable, we constrain the edges so that no cycles are allowed.

To summarize, our graph describes how multiple agents can collaborate to generate a response, where nodes represent agents and a final aggregation of their responses, and edges the interactions between them. %
Correspondingly, the goal of our AQA framework is to learn the most suitable graph configuration for each incoming question.

\begin{figure*}[tp]
\centering
\includegraphics[width=\textwidth]{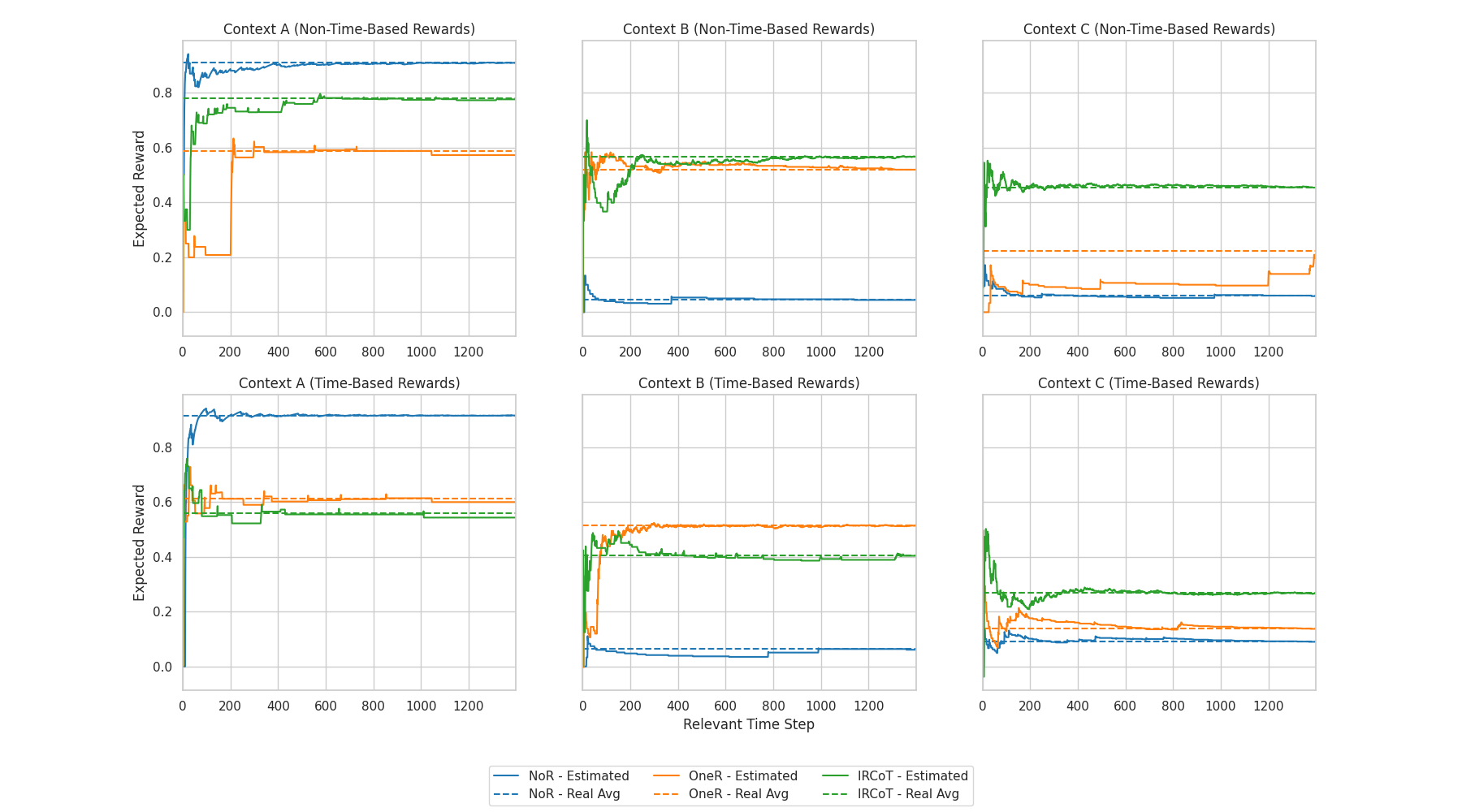}
\caption{LinUCB expected rewards for individual agents action Space, dashed lines depict the real rewards per action.}
\label{fig1}
\end{figure*}

\subsection{Action Space} \label{subsec:actionspace}
Our action space is the set of all possible answer strategies, each represented by a different graph $G$.
Therefore, selecting an answer strategy can be reformulated  as choosing the edges $\mathbf{E}$ of the graph.
\citet{DBLP:journals/corr/abs-2402-16823} apply a policy gradient approach with a graph sampling method to this problem; we differ by taking a CMAB approach.

Our motivation for this difference stems from the following two observations:
(i) the number of agents and their possible interactions is generally limited; and
(ii) many possible graphs have equivalent execution patterns.
Therefore, the set of graphs with unique answering strategies is finite and often of a manageable size for CMAB algorithms.
This is beneficial, since it avoids the high variance of policy gradients and their associated sampling, which can make CMABs more effective. %

In order to apply the CMAB algorithm, we first need to compute the set of possible actions.
Luckily, this only has to be done once, as answering strategies are not question specific.
The set of graphs that represent unique answering strategies are defined by any edges $\mathbf{E}$ that meet the following constraints:
(i) the resulting graph is a directed acyclic graph (DAG);
(ii) the final decision node has at least one incoming edge and no outgoing edges; and
(iii) every agent node either has no edges or there exists a path from it to the final node.
We note that the final constraint prevents islands of interacting agents that do not affect the final response.
The set of all graphs that meet these constraints is defined as $\mathcal{G}$; for the CMAB this is simply the set of possible discrete actions.

\begin{figure*}[t]
\centering
\includegraphics[width=\textwidth]{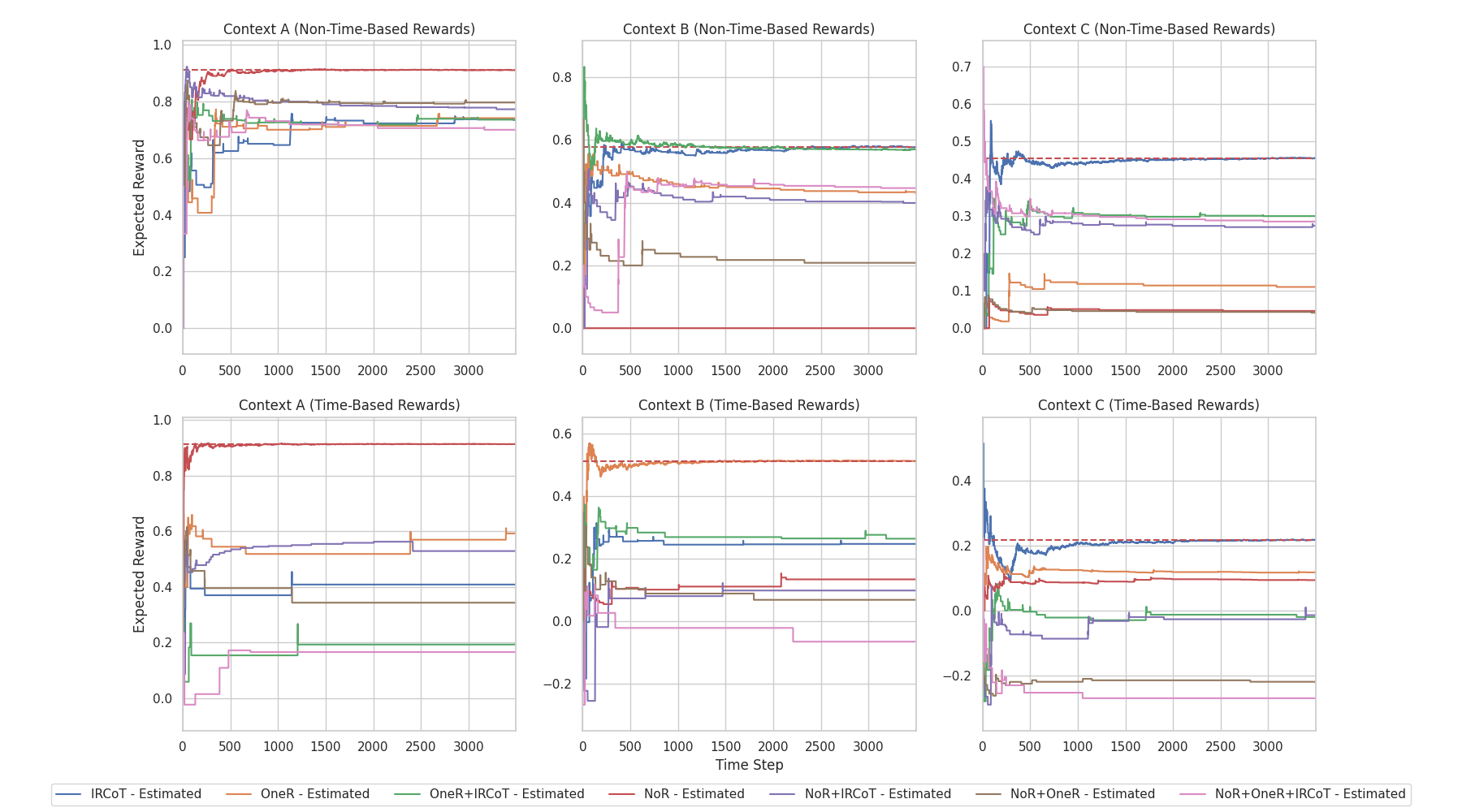}
\caption{LinUCB expected rewards for the collaborative action space, dashed line depicts real reward for the optimal action.}
\label{fig3}
\end{figure*}

\subsection{\textbf{AQA}: Adaptive QA via Contextual Bandit for Multi-Agents Orchestration}
Finally, we describe our complete AQA approach analogous to Algorithm~\ref{alg:linucb}.
We start with the set of agents $\mathbf{L}$ (Line~\ref{alg:line:setofagents}), and subsequently, pre-compute 
the set of possible actions $\mathcal{G}$: every graph that meets the conditions in Section~\ref{subsec:actionspace} (Line~\ref{alg:line:startgraphconstruction}--\ref{alg:line:endgraphconstruction}).
As a result, $\mathcal{G}$ now represents every unique answering strategy possible through the collaboration between the agents in $\mathbf{L}$.
For the CMAB algorithm, $\mathcal{G}$ contains the finite set of discrete actions for the setting.
We note that AQA allows for more constraints on the graphs, e.g., on what agents can interact directly, or a maximum on the number of agents involved to bound computational costs.

Next, we define our objective to be optimized.
As stated before, we wish to balance the performance of a strategy with the costs of its execution, on a per-question level.
For a question $Q_t$ at timestep $t$, let $P_t$ denote performance, i.e., the \emph{correctness} of the generated answer; and let $T_t$ denote the cost for generating it.
Our reward signal is a linear combination of $P_t$ and $T_t$ with parameters $\beta \in [0,1]$:
\begin{equation}\label{eq:reward}
r_t = \beta \cdot P_t - (1-\beta) \cdot T_t.
\end{equation}

\noindent%
Our choice of CMAB algorithm is LinUCB~\citep{li2010contextual};
with the pre-computed action space, our problem matches the LinUCB setting.
Accordingly, we initialize LinUCBs model parameters $\mathbf{A}_a$ and $\mathbf{b}_a$ (Line~\ref{alg:line:startUCBinitialization}--\ref{alg:line:endUCBinitialization}). %
During training, $\mathbf{A}_a$ aims to capture the covariance of context vectors $x_t$, i.e., the correlation between different features, and $\mathbf{b}_a$ the linear relationship between the features and rewards received.
The algorithm enters an indefinite loop (Line~\ref{alg:line:startmainloop}--\ref{alg:line:endmainloop}), that repeats the following process:
First, an incoming question $Q_t$ is received and its features $x_t$ are observed (Line~\ref{alg:line:observequery}).
A context-specific upper bound on the reward for each action is computed and the action with the highest value is selected (Line~\ref{alg:line:startactionselection}--\ref{alg:line:endactionselection}).
The corresponding answering strategy is executed; the generated answer is presented to the user and the performance $P_t$ is observed (Line~\ref{alg:line:observeperformance}).
The reward is computed as a weighted average of $P_t$ and the generation costs $T_t$ (Line~\ref{alg:line:computereward}).
LinUCB updates its parameters accordingly (Line~\ref{alg:line:startupdateparams}--\ref{alg:line:endupdateparams}) and the loop repeats.
Over time, the parameters will converge such that the upper bounds of the best answering strategies are greater than those of other actions.
Consequently, LinUCB can recognize and execute the best answering strategy for every $x_t$, given enough training time.
Importantly, because the answering strategies are selected according to $x_t$, AQA dynamically adapts its strategies to the properties of each individual question.

\begin{figure*}[t]
\centering
\includegraphics[width=\textwidth]{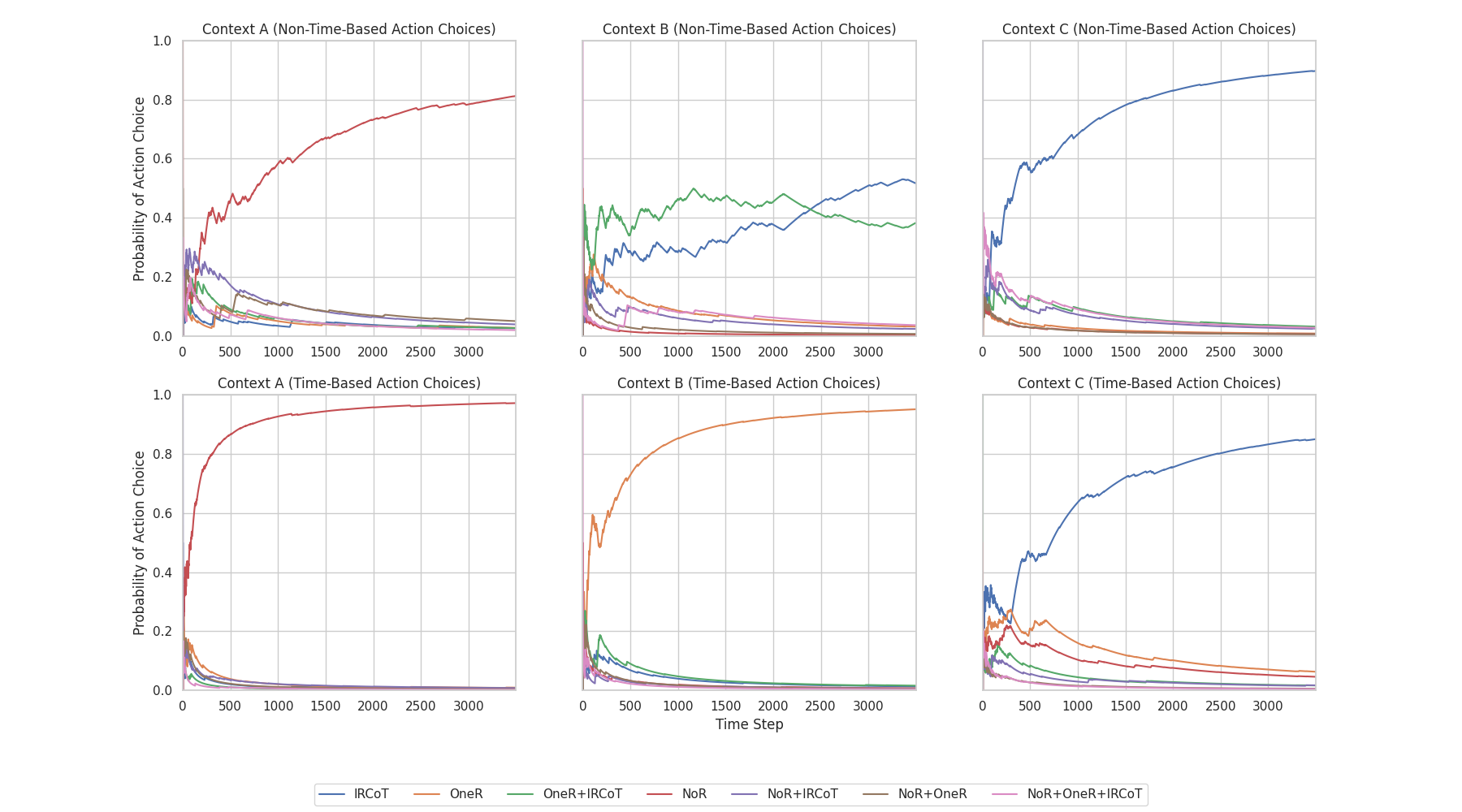}
\caption{LinUCB action selection distribution for the collaborative action space.}
\label{fig4}
\end{figure*}

\section{Experiments}
\subsection{Agents}
For all our experiments, the base LLM of all agents is Flan-T5-XL which is also the model incorporated by \citet{DBLP:conf/acl/TrivediBKS23} and \citet{DBLP:journals/corr/abs-2403-14403}. 
Our implementation of the agents mirrors the exact setup, roles, prompts, retrieval models, and parameters used by \citet{DBLP:journals/corr/abs-2403-14403}.
\subsection{Dataset}
We use the dataset developed by \citet{DBLP:journals/corr/abs-2403-14403} for building and evaluating RAG for QA systems.
It features a diverse array of questions, categorized into varying levels of complexity.
Each question has one of three complexity labels that were generated using Flan-T5-XL: %
Label A indicates that the question can be answered directly without retrieval;
label B requires single-step retrieval;
and label C necessitates multiple reasoning steps along with retrieval. %

For our study, we randomly selected 210 questions and their labels from this dataset for training purposes and 51 questions for testing.
We maintained an equal distribution of the three complexity labels in the sets to ensure a balanced representation for training a CMAB.
The complexity labels are used as the contextual features $x_t$ for the CMAB. %

\subsection{Action Spaces}
We perform two sets of experiments:
\begin{itemize}[leftmargin=*]
\item \textbf{Individual agents action space.}
Our first set covers a simpler setting where graphs are limited to a single edge.
Consequently, the action space represents the responses of individual agents, mimicking the setup of \citet{DBLP:journals/corr/abs-2403-14403}.
We use this setting to verify whether the CMAB can learn the optimal context-agent mapping, without any orchestration.

\item \textbf{Collaborative agents action space.}
The second set of experiments places no additional limitations on the allowed graph structures.
This allows us to evaluate how well the CMAB can optimize when collaboration and dynamic orchestration. 
\end{itemize}

\begin{table}[t]
\centering
\small
\setlength{\tabcolsep}{3pt} %
\begin{tabular}{l cccccc}
\toprule
 & \multicolumn{2}{c}{\textbf{NoR}} & \multicolumn{2}{c}{\textbf{OneR}} & \multicolumn{2}{c}{\textbf{IRCoT}} \\
\cmidrule(lr){2-3} \cmidrule(lr){4-5} \cmidrule(lr){6-7}
 & {F1} & {Time (s)} & {F1}  & {Time (s)}  & {F1}  & {Time (s)} \\
\midrule
\textbf{Context A} & 0.914 & 0.66 & 0.677 & 6.46 & 0.730 & 189.78  \\
\textbf{Context B} & 0.061 & 0.66 & 0.518 & 7.34 & 0.580 & 192.30  \\
\textbf{Context C} & 0.066 & 0.67 & 0.146 & 6.41 & 0.458 & 184.85  \\
\textbf{Overall} & 0.347 & 0.66 & 0.447 & 6.74 & 0.589 & 188.97  \\
\bottomrule
\end{tabular}
\caption{Performance of each individual agent on the training set (by F1-score and average time in seconds).}
\label{tab:train_eval}
\end{table}

\subsection{Setup and Hyperparameters}
We train the LinUCB model for 20 and 50 epochs for the \textit{individual} and \textit{collaborative} action spaces respectively, with exploration parameter $\alpha=2$.
For the reward, we set $\beta=0.5$ and base the computation costs on execution time.
We add the multiplicative penalties: with $S_t$ as execution time in seconds, for the \textit{individual} setting we set $T_t = S_t \frac{\mathbb{I}[S_t > 1]}{1000}$; and for the \textit{collaborative} $T_t = S_t \big(\frac{\mathbb{I}[1 < S_t \leq 10]}{10000} +  \frac{\mathbb{I}[S_t > 10]}{50}\big)$.
Each of our models (different setups of AQA and GPTSwarm) are trained once by setting a fixed random seed. The experiments are ran on a machine with two NVIDIA GeForce RTX 3090 GPUs, each with 24 GB of memory. Computational resources are primarily used for hosting the LLMs and not for training the CB models.

\subsection{Evaluation Metric}
In all experiments, we measure performance ($P_t$) using the F1-score, calculated by comparing each agent's generated response to the gold standard answer for the question.
For computation costs, we report the average execution time per query, 
indicating the time it took for the executed action (agent or collaborative graph) to generate the final response.

\section{Experimental Results}
To be able to interpret the results of our CMAB experiments, we first assess the individual performance of each agent defined in $L'$ on the training dataset (shown in Table \ref{tab:train_eval}). This evaluation reveals distinct performance levels for each agent across different complexity levels.
The NoR agent excels in the simplest scenarios, while IRCoT surpasses all others at complexity levels B and C. At level B, IRCoT slightly outperforms OneR (.580 versus .518), albeit at a significantly higher time cost (192.30 versus 6.41 seconds). 
This increase is due to IRCoT's potential for up to 10 rounds of retrieval and CoT, potentially enhancing document relevance and ``reasoning" toward the right answer, but at the expense of greater response time in practical applications. 
In real-world, the time cost translates to the waiting time for the user to get a potential correct response from the system.

\subsection{Optimization of Individual Agents Setup}
In this set of experiments, we aim to verify whether CMAB can learn to map the complexity level of incoming questions to the most optimal action in the individual action space (where each action consists of choosing among possible set of agents $L$). To test this hypothesis, we need to verify whether training LinUCB causes the expected reward for the best action to converge toward the actual reward for that action.
We perform our experiments in both time-agnostic and time-based modes to assess the effect of time consideration  in our framework:
\begin{itemize}[leftmargin=*]
\item \textbf{Time-agnostic reward.}
The first row of Figure~\ref{fig1} depicts the estimated reward for time-agnostic optimization progress. Looking at this figure, we can see the results of Table \ref{tab:train_eval} also depicted here, as the CMAB learns to choose the best-performing action for each relevant context (NoR for context label A and IRCoT for context labels B and C).
\item \textbf{Time-based reward.} 
Incorporating the time cost into the reward calculation, we observe from the second row of Figure \ref{fig1} that the model progressively selects efficient actions for context B. Specifically, it favors the OneR agent, which balances robust performance with reasonable processing time.
\end{itemize}

\begin{figure}[tp]
\centering
\begin{minipage}{0.5\columnwidth}
    \centering
    \includegraphics[width=\textwidth]{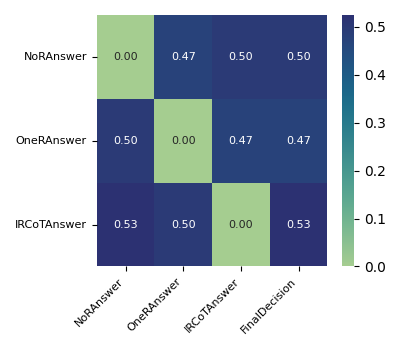}
\end{minipage}\hfill 
\begin{minipage}{0.5\columnwidth}
    \centering
    \includegraphics[width=\textwidth]
    {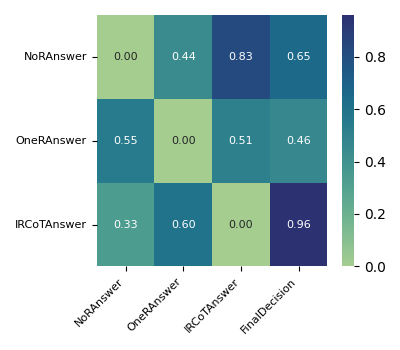}
\end{minipage}
\caption{Edge probabilities distribution among NoR, OneR, IRCoT, and FinalDecision nodes before (left) and after (right) optimization using GPTSwarm. \citep{DBLP:journals/corr/abs-2402-16823}.
}
\label{fig:both_optimizations}
\end{figure}

\subsection{Optimization of Collaborative Agents Setup}
In this step, our goal is to verify whether AQA is able to learn an optimal mapping between the complexity level of questions and the suitable collaborative graph configuration.%

\begin{itemize}[leftmargin=*]
\item \textbf{Time-agnostic reward.}
In Figures \ref{fig3} and \ref{fig4}, the top rows depict the expected rewards and action selection distribution for LinUCB, without considering time in the reward. These results reveal optimal actions identified for each complexity label; NoR for context A, and IRCoT for contexts B and C, based on F1-scores. The dotted lines in the figures depict the real reward calculated for the best possible action per context.

\item \textbf{Time-based reward.} 
The bottom rows of Figures \ref{fig3} and~\ref{fig4} depict results for training while including time cost in the reward calculation. Considering both performance and the time cost, the real reward for best actions per labels are shown in the plots as dotted red lines. Including time in reward calculation, we observe that the model preferentially selects actions that balance efficacy with reasonable execution times. Specifically, in context B, it prioritizes the configuration with OneR linked to the $v_\text{final}$ over the configuration with IRCoT or a combination of better-performing agents (i.e., OneR+IRCoT) connected to $v_\text{final}$.
\end{itemize}

\noindent%
The results presented above demonstrate that our proposed setup is useful in adaptive orchestration in both a disjoint LLM setup, where the task is to choose the best agent for a specific task, but also in an orchestration setup where multiple nodes/modules might be involved to answer a question and the task there is to choose the best communication configuration between these modules considering the characteristics of the incoming question.

\subsection{AQA vs GPTSwarm for Orchestration}

Here, we aim to assess how a non-adaptive orchestration optimization affects the performance. To this aim, we compare AQA and GPTSwarm \citep{DBLP:journals/corr/abs-2402-16823} in collaborative mode, train them both, and assess them on our test set. 
Both models are trained and tested using the same graph structure described in Section~\ref{subsection:collab}. As GPTSwarm does not allow for the inclusion of either context or time cost in the proposed framework, in our setup (similar to what \citet{DBLP:journals/corr/abs-2402-16823} do in their MMLU experiment) at both training and test time we feed each question to the framework and the optimization gradually converges to the \textit{most optimal} graph configuration by optimizing towards better performance (F1-score).

The edge probability distribution of the graph before and after REINFORCE optimization is shown in Figure \ref{fig:both_optimizations}. As we can see, after training for 200 epochs, the final optimized graph consists of both NoR and IRCoT agents, and the OneR agent's edge to the FinalDecision node is pruned out (edges with probabilities lower than .5 are pruned out based on the implementation by \citet{DBLP:journals/corr/abs-2402-16823}). This shows, as expected, that this static framework fails to adapt to different complexity levels by routing for a good enough answering strategy that is also time-efficient, as it only favors higher accumulative F1 scores across time. 

The evaluation results on the test set are shown in Table~\ref{tab:test_eval}. The first two columns show the CMAB model (LinUCB) with and without time considered in the reward implementation respectively and the last column is the final optimized graph using REINFORCE. 
Using a fixed orchestration by GPTSwarm, we observe a fall in performance for different complexity levels, as it is not possible to adapt the strategy based on the characteristics of the question, which leads to a lower overall performance compared to AQA (NT and T). Also, as the fixed configuration in GPTSwarm is also more sophisticated (i.e. both NoR and IRCoT have edges to the final decision node), the time cost is constantly higher compared to AQA.

\begin{table}[t]
\centering
\small
\setlength{\tabcolsep}{3pt} %
\begin{tabular}{l cc cc cc}
\toprule
 & \multicolumn{2}{c}{\textbf{AQA (NT)}} & \multicolumn{2}{c}{\textbf{AQA (T)}} & \multicolumn{2}{c}{\textbf{GPTSwarm}} \\
\cmidrule(lr){2-3} \cmidrule(lr){4-5} \cmidrule(lr){6-7}
 & {F1} & {Time} & {F1} & {Time} & {F1}  & {Time} \\
\midrule
\textbf{Context A} & 1.0\phantom{00} & \phantom{0}6.18 & 1.0\phantom{00} & \phantom{0}6.18 & 0.862 & 12.78 \\
\textbf{Context B} & 0.568 & 12.04 & 0.539 & \phantom{0}8.73 & 0.327 & 12.79 \\
\textbf{Context C} & 0.523 & 11.75 & 0.523 & 11.75 & 0.317 & 12.76 \\
\textbf{Overall} & 0.697 & \phantom{0}9.99 & 0.687 & \phantom{0}8.89 & 0.502 & 12.78 \\
\bottomrule
\end{tabular}
\caption{Evaluation of AQA (NT: Time-agnostic reward, T: Time-based reward) and GPTSwarm \citep{DBLP:journals/corr/abs-2402-16823} on the test set by F1-score and time (log-transformed, in ms). Both have been trained on the training set. For GPTSwarm, the final optimized graph configuration is used for evaluation. 
}
\label{tab:test_eval}
\end{table}

\section{Discussion and Conclusion}
In this study, we introduced \textbf{AQA}, a novel adaptive QA framework that uniquely frames the adaptive QA problem as a contextual multi-armed bandit problem, where the action space is a set of  graph structures among LLM agents that describe their interactions.
Thereby, the AQA framework dynamically orchestrates the collaboration of multiple agents in response to specific question characteristics. 
We evaluated our approach using a multi-complexity-level QA dataset and validated that our approach can successfully recognize the optimal mapping for each question type.

Future work could consider integrating additional models into the graph component and employing more advanced contextual features for the questions, such as semantic representations or predefined textual features.
This direction could also explore more advanced CB methods than the LinUCB to enhance the scalability and applicability of our framework. 
Another avenue for the future work is the comparison of our setup which is based on the society of mind (SoM) theory \citep{minsky1988society, DBLP:journals/corr/abs-2305-17066}, to mixture of experts (MoE) based setups \citep{zhou2022mixture, DBLP:journals/corr/abs-2401-04088} as there are clear similarities but also distinctions such as the agent roles definition, the routing strategy, and the decision space. In SoM, each agent role is predefined and no specific learning is happening at the agent (LLM) level. Also, SoM allows for explicit natural language-based communication between agents, which is not the case in MoE architectures. It is interesting to compare the two setups to understand how they potentially outperform or complement each other for future LLM-based system designs. 

\section*{Resources}
The code and other resources needed to reproduce the results of this paper are publicly available at \url{https://github.com/informagi/AQA}.

\section*{Acknowledgments}
This research was supported by the Dutch Research Council (NWO), under project numbers VI.Veni.222.269, 024.004.022, NWA.1389.20.183, and KICH3.LTP.20.006, and the EU’s Horizon Europe program under grant No 101070212. All content represents the opinion of the authors, which is not necessarily shared or endorsed by their respective employers and/or sponsors. 

We would also like to thank Nik Vaessen for the useful discussions that helped inform aspects of this work.

\bibliography{references}

\end{document}